\documentclass[sigconf,screen]{acmart}

\usepackage{balance} 
\usepackage{wrapfig}
\usepackage{comment}
\usepackage{colortbl}
\usepackage[english]{babel}
\usepackage{hyperref}

\usepackage{dblfloatfix} 

\usepackage{amsthm} 

\usepackage{color,soul}
\usepackage{graphics}
\usepackage{hyperref}
\usepackage{lipsum}
\usepackage{tikz}
\usepackage{enumitem}	
\usepackage{listings} 
\usepackage{booktabs}	
\usepackage{lipsum}

\usepackage{capt-of}	
\usepackage{diagbox}
\usepackage{multirow}
\usepackage{todonotes}  
\usepackage{subcaption} 

\definecolor{refkey}{rgb}{249,158,26}
\definecolor{labelkey}{rgb}{0,1,0}
\definecolor{airforceblue}{rgb}{0.36, 0.54, 0.66}
\definecolor{applegreen}{rgb}{0.55, 0.71, 0.0}

\usepackage{mathrsfs}	
\usepackage{amsfonts}

\usepackage[export]{adjustbox} 

\usepackage{algorithm,algorithmic,amsmath} 
\usepackage{boldline}					

\usepackage{multirow}

\newtheorem*{definition}{Definition}

\definecolor{frenzyorange}{RGB}{249, 158, 26}
\newcommand*\circled[1]{\tikz[baseline=(char.base)]{
			\node[shape=circle,draw,text=black,inner sep=1pt] (char) {#1};}}
		
\renewcommand{\paragraph}[1]{\vskip 3pt\noindent\textbf{#1 }}	 

  {\begin{list}{$\bullet$}%
     {\setlength{\parsep}{0pt}%
      \setlength{\topsep}{0pt}%
      \setlength{\itemsep}{2pt}}}%
  {\end{list}}
\newcommand\Note[1]{\sethlcolor{applegreen} \hl{#1}} 
\newcommand\Noted[1]{} 

\newcommand\xzlNote[1]{\sethlcolor{yellow} \hl{#1}} 

\definecolor{darkblue}{rgb}{0.0, 0.0, 0.55}
\definecolor{mygreen}{HTML}{ADFF2F}
\definecolor{mylightgray}{gray}{0.8}

\newenvironment{myitemize}%
  {\begin{itemize}
	[leftmargin=0cm,
		itemindent=.3cm,
		labelwidth=\itemindent,
		labelsep=0pt,
		parsep=1pt,
		topsep=1pt,
		itemsep=1pt,
		align=left]
  }%
  {\end{itemize}}    

\newenvironment{myenumerate}%
  {\begin{enumerate}
	[leftmargin=.cm,itemindent=.5cm,labelwidth=\itemindent,
		labelsep=0pt,
		parsep=1pt,
		topsep=1pt,
		itemsep=3pt,
		align=left]
  }%
  {\end{enumerate}}    

\newcommand\sect[1]{Section~\ref{sec:#1}}	

\newcommand{\code}[1]{\texttt{\small{#1}}}	

\newcommand{\sys}{STI}
\newcommand{\tfm}{transformer}
\newcommand{\Tfm}{Transformer}
\newcommand{\bt}{bitwidth}

\makeatletter
\def\@copyrightspace{\relax}
\makeatother

\setcopyright{rightsretained}
\acmPrice{}
\acmDOI{10.1145/3575693.3575698}
\acmYear{2023}
\copyrightyear{2023}
\acmSubmissionID{asplosb23main-p32-p}
\acmISBN{978-1-4503-9916-6/23/03}
\acmConference[ASPLOS '23]{Proceedings of the 28th ACM International Conference on Architectural Support for Programming Languages and Operating Systems, Volume 2}{March 25--29, 2023}{Vancouver, BC, Canada}
\acmBooktitle{Proceedings of the 28th ACM International Conference on Architectural Support for Programming Languages and Operating Systems, Volume 2 (ASPLOS '23), March 25--29, 2023, Vancouver, BC, Canada}
\received{2022-07-07}
\received[accepted]{2022-09-22}



\begin{document}

\title{STI: Turbocharge NLP Inference at the Edge via Elastic Pipelining}



\author{Liwei Guo}
\affiliation{%
	\institution{University of Virginia}
	\city{}
	\state{}
	\country{USA}
}
\email{lg8sp@virginia.edu}
\author{Wonkyo Choe}
\affiliation{%
	\institution{University of Virginia}
	\city{}
	\state{}
	\country{USA}
}
\email{wonkyochoe@virginia.edu}
\author{Felix Xiaozhu Lin}
\affiliation{%
	\institution{University of Virginia}
	\city{}
	\state{}
	\country{USA}
}
\email{felixlin@virginia.edu}


\date{}

\thispagestyle{empty}

\begin{abstract}


Natural Language Processing (NLP) inference is seeing increasing adoption by mobile applications, where \textit{on-device} inference is desirable for crucially preserving user data privacy and avoiding network roundtrips. 
Yet, the unprecedented size of an NLP model stresses both latency and memory, creating a tension between the two key resources of a mobile device. 
To meet a target latency, holding the whole model in memory launches execution as soon as possible but increases one app's memory footprints by several times, limiting its benefits to only a few inferences before being recycled by mobile memory management.
On the other hand, loading the model from storage on demand incurs IO as long as a few seconds, far exceeding the delay range satisfying to a user;
pipelining layerwise model loading and execution does not hide IO either, due to the high skewness between IO and computation delays.

To this end, we propose Speedy Transformer Inference (STI).
Built on the key idea of maximizing IO/compute resource utilization on the most important parts of a model, \sys{} reconciles the latency v.s. memory tension via two novel techniques.
First, model sharding.
\sys{} manages model parameters as independently tunable \textit{shards}, and profiles their importance to accuracy.
Second, elastic pipeline planning with a preload buffer. 
\sys{} instantiates an IO/compute pipeline and uses a small buffer for preload shards to bootstrap execution without stalling at early stages;
it judiciously selects, tunes, and assembles shards per their importance for resource-elastic execution, maximizing inference accuracy.

Atop two commodity SoCs, we build \sys{} and evaluate it against a wide range of NLP tasks, under a practical range of target latencies, and on both CPU and GPU. 
We demonstrate that \sys{} delivers high accuracies with 1--2 orders of magnitude lower memory, outperforming competitive baselines. 
\end{abstract}

\maketitle


\begin{figure}[h]
	\includegraphics[width=0.48\textwidth]{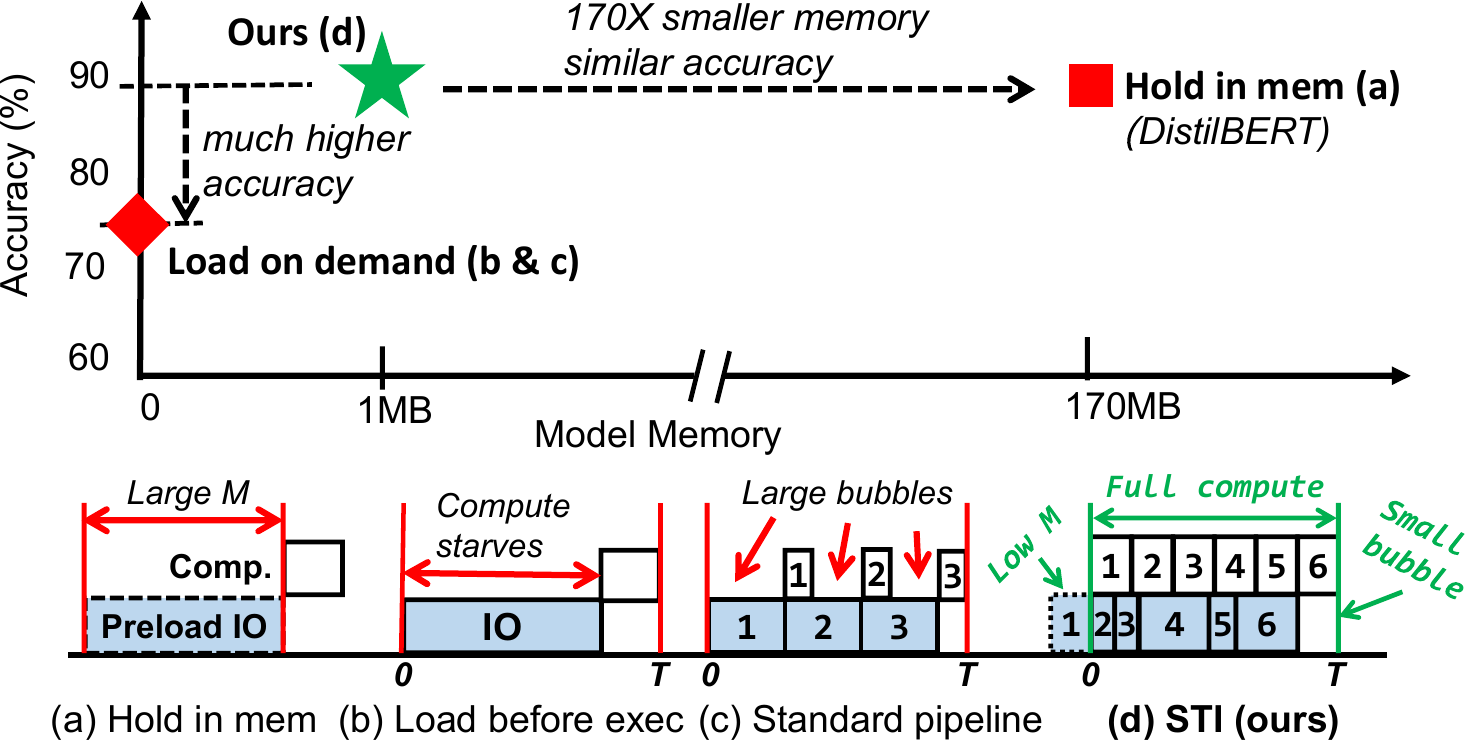}
	\caption{Comparison of model execution methods. Our method achieves high accuracy at low memory cost. T: target latency. M: model memory for Transformer weights.}
	\label{fig:overview-timeline}
\end{figure}

\section{Introduction}
\label{sec:intro}



Natural Language Processing (NLP) is seeing increasing adoption by mobile applications~\cite{ipa-survey}. 
For instance, a note-taking app allows users to verbally query for old notes and dictate new notes. 
Under the hood, the app invokes an NLP model in order to infer on user input.
It is often desirable to execute NLP inference \textit{on device}, which 
crucially preserves user data privacy and eliminates long network trips to the cloud~\cite{edgebert,char-inference-edge}.



NLP inference stresses mobile devices on two aspects. 
(1) Impromptu user engagements. 
Each engagement comprises a few turns~\cite{ipa-study-denmark}; 
users expect short delays of no more than several hundred ms each turn~\cite{mobile-study-chi15}, often mandated as target latencies~\cite{edgebert}. 
(2) Large model size. 
Designed to be over-parameterized~\cite{16heads,lottery-tickets-iclr20},
today's NLP models are hundred MBs each~\cite{bert,distilbert,gpt-2}, much larger than most vision models~\cite{mobilenetv2,shufflenet}.
As a common practice, separate NLP model instances are fine-tuned for tasks and topics, e.g. one instance for sentiment classification~\cite{bert-sentiment} and one for sequence tagging~\cite{2b-not-2b}, 
which further increase the total parameter size on a mobile device. 
How to execute NLP models? 
There are a few common approaches (Figure~\ref{fig:overview-timeline}).
(1) \textit{Hold in memory}: 
preloading a model before user engagement or making the model linger in memory after engagement. 
The efficacy is limited: 
a model in memory increases one app's memory footprint (often less than 100MB~\cite{marvin-atc20,lee-access21}) by a few times, making the app a highly likely victim of the mobile OS's low memory killer~\cite{android-lmkd}; 
as user engagements are bursty and each consists of as few as 1-3 model executions~\cite{ipa-study-denmark}, a lingering model likely benefits no more than 2 executions before its large memory is reclaimed by the OS; 
since user engagements are impromptu~\cite{mobile-usage-imc15,mobile-usage-sigir20}, predicting when to preload/unload models is challenging. 
(2) \textit{Load on demand}.
The problem is the long IO delays for loading a NLP model. 
For instance, DistilBERT, a popular model optimized for mobile, takes 2.1 seconds to load its 170 MB parameters as we measured, far exceeding user desirable latencies of several hundred ms. 
To hide IO delays, one may stream model parameters from storage to memory during computation: execute model layer $k$ while loading parameters for layer $k+1$.
While such an IO/compute pipeline was known in ML~\cite{mengwei-coldstart,hongyu-tinynn}, directly applying it to NLP inference is ineffective: 
the core parts of NLP models such as attention has a skewed IO/compute ratio 
due to low arithmetic intensity~\cite{hanrui20}. 
As a result, most of the time (>72\%) the computation is stalling. 


These approaches suffer from common drawbacks:
(1) key resources -- 
memory for preload and IO/compute for model execution -- are managed in isolation and lack coordination; 
(2) obliviousness to a model's parameter importance, i.e. which parameters matter more to model accuracy. 
Hence, the preload buffer unnecessarily holds parameters that could have been streamed in parallel to execution; 
IO unnecessarily loads parameters that the computation cannot consume within the target latency.
The results are memory waste, frequent pipeline stalls, and inferior model accuracy due to low FLOPs. 


\paragraph{Our design}
We present an engine called \sys{}. 
Addressing the drawbacks above, \sys{}
integrates on-demand model loading with lightweight preload, getting the best of both approaches. 

\noindent \textit{(1) A model as resource-elastic shards.}
%
The engine preprocesses an N-layer model: 
partitioning each layer into $M$ shards; 
compressing each shard as $K$ fidelity versions, 
each version with a different parameter bitwidth. 
The engine therefore stores the $N \times M \times K$ shard versions on flash. 
At run time, the engine assembles a \textit{submodel} of its choice:
a subset of $n$ layers ($n<=N$); 
$m$ shards ($m<=M$) from each selected layer; 
a fidelity version for each selected shard. 
\textit{Any such submodel can yield meaningful inference results}, albeit with different accuracies and resource costs. 
Our model sharding is a new combination of existing ML techniques~\cite{dynabert,gobo}. 

In this way, the engine can dynamically vary a model's total execution time, 
adjust  IO/compute ratios for individual shards, 
and allocate IO bandwidth by prioritizing important shards. 




\noindent 
\textit{(2) Preload shards for warming up pipeline.}
The engine maintains a small buffer of preload shards, adjusting the size to available memory.
Instead of trying to hold the entire model, 
it selectively holds shards from a model's bottom layers (closer to input). 
Upon user engagement, the engine can start executing the early stage of a pipeline with much of the parameters already loaded, which otherwise would have to stall for IO. 

\noindent 
\textit{(3) A joint planner for memory, IO, and computation.}
The engine's planner selects shards and their versions to preload and to execute. 
Its goal is to compose a submodel that simultaneously meets the target latency, minimizes pipeline stalling, and maximizes accuracy. 




Towards this goal, our ideas are (1) set layerwise IO budgets according to layerwise computation delays and (2) allocate IO budgets according to shard importance. 
To plan, \sys{} first decides a submodel that can be computed under the target latency.
The engine then sets \textit{accumulated IO budgets} (AIBs) at each layer to be the computation delays of all prior layers; 
it further treats the available memory for preload shards as \textit{bonus} IO budgets to all layers. 
Having set the budgets, the engine iterates over all shards, allocating extra bitwidths to loading important shards and hence debiting IO budgets of respective layers. 
The engine  preloads the first $k$ shards in the layer order that maximize the usage of preload memory size $|S|$ but not exceeding $|S|$. 

\paragraph{Results}
We implement \sys{} atop PyTorch and demonstrate it on mobile CPU and GPU of two embedded platforms. 
On a diverse set of NLP tasks, \sys{} meets target latencies of a few hundred ms while yielding accuracy comparable to the state of the art. 
We compare \sys{} against competitive baselines enhanced with recent ML techniques~\cite{dynabert,gobo} as illustrated in Figure~\ref{fig:overview-timeline}.
Compared to holding a model in memory, \sys{} reduces parameter memory by 1-2 orders of magnitude to 1--5MB, while only seeing accuracy drop of no more than  0.1 percentage points; 
compared to existing execution pipelines, \sys{} increases accuracy by 5.9-54.1 percentage points as its elastic pipeline maximizes both compute and IO utilization. 

\paragraph{Contributions}
The paper makes the following contributions:
\begin{itemize}
\item Model sharding, 
allowing the engine to fine control an NLP model's total computation time and finetune each shard's IO time according to resource constraints and shard importance. 

\item A pipeline with high IO/compute utilization: a small preload buffer for warming up the pipeline; 
elastic IO and computation jointly tuned to minimize pipeline bubbles and maximize model accuracy. 

\item A two-stage planner for the pipeline: 
picking a submodel, tracking layerwise IO budgets, and prioritizing importance shards in resource allocation. 
\end{itemize}


\section{Motivations}
\label{sec:motiv}


\subsection{\Tfm{} on Mobile Devices}
\label{sec:motiv:primer}
%

\begin{figure}[t]
	\vspace{1em}
	\centering
	\includegraphics[width=0.38\textwidth]{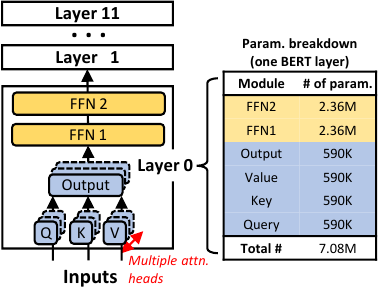}

	\caption{(Left) The BERT model comprising transformer layers and (Right) the number of 32-bit floating point parameters within a layer~\cite{transformer}.}
	\label{fig:motiv:transformer}
\end{figure}

\paragraph{A primer on \tfm{}}
Figure~\ref{fig:motiv:transformer}
shows the architecture of Transformer~\cite{transformer}, the modern NN developed for NLP tasks.  
Compared with traditional NNs (e.g. LSTM~\cite{lstm}), it features a unique Multi-Headed Attention (MHA) mechanism. 
MHA extracts features at sequence dimension by modeling pairwise word interactions through many \textit{attention heads} (typically 12), which are backed by three fully-connected (i.e. linear) layers, namely Query (Q), Key (K), Value (V).
Given an input, each attention head independently contributes an attention score as one representation of the feature space.
Scores across attention heads are concatenated via a linear output layer (O) and then projected into higher feature dimensions by two linear layers in the point-wise Feed-Forward Network (FFN) module. 

Due to the large number of fully connected layers, a \tfm{} based model contains over 100 million parameters.
As a result, a typical pretrained model is of a few hundred MBs.
For instance, BERT~\cite{bert} as one of the most popular model is over 400MB large.


\paragraph{Resource demands}
(1) \textit{Low latencies}. 
Prior studies show that users expect mobile devices to respond in several hundred milliseconds, and their satisfaction quickly drops as latency grows beyond around 400ms~\cite{nlp-delay-study}.  
(2) \textit{Large model parameters}. 
The scale of NLP parameters is unprecedented for on-device machine learning.
Even DistilBERT~\cite{distilbert} optimized for mobile has nearly 200MB of parameters, contrasting to popular vision models which are as small as a few MBs~\cite{mobilenetv2,shufflenet}. 
Such numerous parameters stress both memory capacity and IO for loading them. 

Besides parameters, model execution also allocates memory for intermediate results. 
Yet, such data has short lifespans and does entail loading from storage.
Hence, it can be served with a relatively small working buffer sufficient to hold a model tile (often a few MBs); the size does not grow with the model size. 
We therefore do not optimize for it.

\subsection{\Tfm{}s Challenge Existing Paradigms}
\label{sec:motiv:char}


Existing paradigms are inadequate, as shown in Figure~\ref{fig:overview-timeline}.

\paragraph{\textit{First, hold in memory.}}
An app may keep model files lingering in memory or even \textit{pin} them; 
thus, the model can start execution anytime without IO delays. 
For how long the app holds the model depends on its prediction of future user engagements. 

The major drawback is that an in-memory model will take hundreds of MBs of memory, bloating an app's memory footprint which is often less than 100 MBs~\cite{marvin-atc20,lee-access21}.
When an app's memory footprint is much larger than its peers, it becomes a highly likely victim of mobile memory management, which aggressively kills memory-hungry apps~\cite{marvin-atc20}. 
Once killed, the app has to reload the model for the next engagement.
Furthermore, precise prediction of user engagement is difficult, as mobile apps often exhibit sporadic and ad hoc usage~\cite{livelab,mobile-study-chi15}. 
To exacerbate the problem, co-running apps may invoke separate models for their respective tasks, e.g. for sentiment analysis and for next-word prediction. 



\paragraph{\textit{Second, load before execute.}}
As the default approach by popular ML frameworks~\cite{pytorch-web,tensorflow-web}: 
upon user engagement, the app sequentially loads the model and executes it. 
As we measured on a modern hexa-core Arm board (see Table~\ref{tab:platform}), it takes 3.6 seconds to execute DistilBERT, among which 3.1 seconds are for loading the whole 240 MB model file.  
Prior work observed similar symptoms of slow start of model inference~\cite{dnn-mobile-lat-var,mengwei-coldstart}.

\paragraph{\textit{Third, pipelined load/execution.}}
To hide IO delays, 
one may leverage layerwise execution of ML models~\cite{gpipe,pipedream} and overlap the layer loading IO and execution~\cite{mengwei-coldstart,hongyu-tinynn}.
This approach is barely effective for on-device NLP due to the high skewness between IO delays and computation delays. 
As we measured, a layer in DistilBERT requires 339 ms for parameter load while only 95 ms to compute.
The root causes are (1) low arithmetic intensity in Transformer's attention modules~\cite{patiDemystifyingBERTImplications2021} and (2) mobile device's efficiency-optimized flash, 
which limits the rate of streaming parameters from storage to memory. 
As a result, the pipeline is filled with bubbles and the computation stalls most of the time at each model layer. 

\sect{eval} will compare our system against these approaches.

\subsection{Model Compression Is Inadequate}

For efficient NLP inference, 
a popular category of techniques is model compression, including pruning networks (e.g. layers~\cite{distilbert} and attention heads~\cite{head-prune-acl19}), reducing feature dimensions~\cite{mobilebert}, and sharing weights across layers~\cite{albert}.
A notable example is DistilBERT~\cite{distilbert}: 
through distilling knowledge, it prunes half of BERT's layers, shrinking the model by $2\times$. 

Still, model compression \textit{alone} is inadequate. 
(1) While one may compress a model to be sufficiently small \sloppy{(e.g. $\sim$10MBs~\cite{bibert})} so that the load delay or the memory footprint is no longer a concern, the resultant accuracy is inferior, often unusable~\cite{efficient-transformer-survey}.
(2) The execution pipeline's bubbles still exist: compression often scales model compute and parameters \textit{in tandem}, without correcting the computation/IO skewness. 
Hence, compute is still being wasted.
(3) Most compression schemes lack flexibility as needed to accommodate diverse mobile CPU, GPU, and IO speeds. 
They either fix a compression ratio or require model re-training to adjust the ratios, which must done by the cloud for each mobile device. 

\sect{eval} will evaluate the impact of model compression.

\section{Design overview}

%
%
%
%

\subsection{The System Model}

\sys{} incarnates as a library linked to individual apps. 
For complete NLP experience, we assume that the app incorporates other components such as automatic speech recognition (ASR), word embedding, and speech synthesis~\cite{shi2021emformer,recssd, flashembedding, bandana}.
As they often run much faster than model execution and are orthogonal to \sys{},
this paper does not optimize for them. 


\sys{} loads and executes a model by layer: 
it loads one layer (comprising multiple shards) as a single IO job, decompresses all the shards in memory, and computes with the layer as a single compute job. IO and compute jobs of different layers can overlap. 
\sys{} does not use smaller grains (e.g. load/execute each \textit{shard}) as they leave the IO and GPU bandwidth underutilized, resulting in inferior performance. 

\sys{} allocates two types of memory buffers.

 \begin{myitemize}
 	\item 
 	\textit{Preload buffer} holds shards preloaded selectively.
	\sys{} keeps the buffer as long as the app is alive. 
	\sys{} can dynamically change the buffer size as demanded by the app or the OS.
 	
 	
 	\item 
 	\textit{Working buffer} holds a layer's worth of intermediate results and uncompressed parameters. 
 	The buffer is temporary, allocated before each execution and freed afterward. The buffer size is largely constant, not growing with the model size; 
 	it is not a focus of \sys{}. 
 \end{myitemize}

%
%
%
%
%



\begin{figure}[t]
	\centering
	\includegraphics[width=0.45\textwidth]{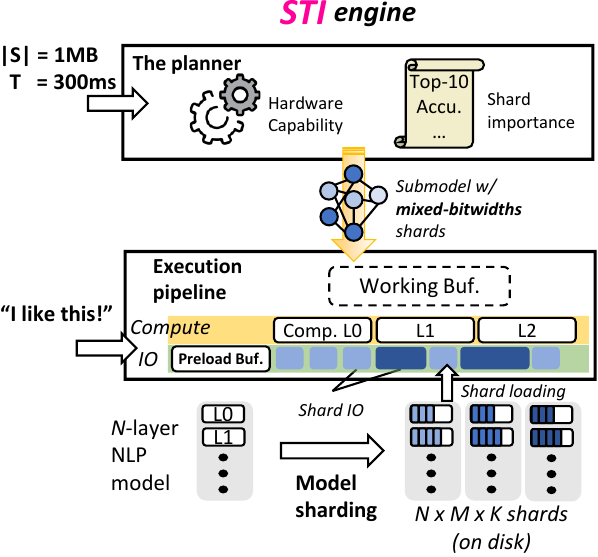}
	\caption{System architecture of \textbf{S}peedy \textbf{T}ransformer \textbf{I}nference (\sys{}) and workflow.}
	\label{fig:design:overview}
\end{figure}

\subsection{The Operation}

The \sys{} architecture is shown in Figure~\ref{fig:design:overview}. 
\sys{} preprocesses a given language model (e.g. DistilBERT finetuned for sentiment analysis): 
decomposing the model into shards and profiling shard importance (\sect{planning}). 
As a one-time, per-model effort, 
the preprocessing is expected to be done in the cloud prior to model deployment to mobile devices; 
as preprocessing only requires lightweight model transformation (as opposed to expensive re-training~\cite{roberta}), it can be done on device as needed. 
The resultant model shards are stored alongside apps. 

\sys{} profiles each device's hardware once. 
The goal is to measure IO and computation delays in executing a language model; the profiling results serve as the basis for pipeline planning.
To do so, \sys{} loads and executes a Transformer layer in different bitwidths. 

As an app launches, \sys{} is initialized as part of the app. 
The app specifies which NLP model(s) it expects to execute, as well as the corresponding target latencies $T$s and preload buffer sizes $|S|$s. 
Later, the app can update $T$s and $|S|$s at any time. 
For each expected model, \sys{} plans a separate execution pipeline with  separate preload model shards. 
\sys{} plans a pipeline once and executes it repeatedly. 
Replanning is necessary only when a model's $T$ or $|S|$ is changed by the app or OS. 



Upon user engagement, \sys{} executes a pipeline for the requested model. 
Since planning is already done beforehand, \sys{} simply loads and executes the shards that have been selected in planning.

\subsection{Example Execution Scenarios}

\paragraph{One-shot execution}
In this scenario, 
a user engagement consists of one turn, executing the model once. 
With preloaded shards, \sys{} executes the pipeline without stalling in bottom layers, which are close to input. 
\sys{} uses the working buffer during the execution and frees it right after. 
Throughout the execution, the content of preload buffer is unchanged. 

\paragraph{A few back-to-back executions}
One engagement may comprise multiple executions (often no more than 3)~\cite{ipa-study-denmark}.
The scenario is similar to the above, except for the opportunity of caching already loaded shards between executions. 
To this end, the app may request to enlarge the preload buffer so it selectively caches the loaded shards. 
In subsequent executions, \sys{} no longer reloads these shards; 
its planner redistributes the freed IO bandwidth to other shards (\sect{planning}), loading their higher-fidelity versions for better accuracy. 
After the series of executions, the app may choose to keep the additional cached shards as permitted by the OS or simply discard them.


\subsection{Applicability}
\sys{} supports Transformer-based models~\cite{hat,dynabert,roberta}. 
This paper focuses on classification tasks (BERT and its variants), which underpin today's on-device NLP. 
Although \sys{}'s key ideas apply to \textit{generative} models such as GPT-2~\cite{gpt-2}, their wide adoption on mobile (in lieu of template-based responses~\cite{mctear2016conversational}) is yet to be seen; we consider them as future work.


\sys{} keeps a model's execution time under a target latency $T$. 
However, it \textit{alone} is insufficient to keep the \textit{total wall-clock time} under $T$. 
Such a guarantee would require additional OS support, e.g. real-time scheduling. 
\sys{} lays the foundation for such a guarantee. 


\sys{} expects a small preload buffer.
It can, however, work without such a buffer (i.e. ``cold start'' every time), 
for which its elastic sharding and pipeline still offer significant benefits as we will show in \sect{eval}. 

On future hardware/workloads, we expect \sys{}'s benefit to be more pronounced: 
mobile compute continues to scale (due to advances in technology nodes and  accelerators); 
users expect results in higher accuracy; 
NLP models are becoming larger. 
All these lead to higher computation/IO skewness, necessitating an elastic pipeline of loading and execution.

\section{Elastic model sharding}
\label{sec:design:sharding}

%
%



\subsection{Key Challenges}
We solve a key challenge: how to partition the model into individual shards? 
Set to enable the resource elasticity of a model (i.e. depths/widths/fidelity), the shards must meet the following criteria: 
\begin{myitemize}	
	\item 
	\textit{Elastic execution.}
	Shards must preserve the same expressiveness of the attention mechanism and can execute partially to produce meaningful results. 
	
	\item
	\textit{Tunable IO.} 
	The IO delays of shards must be tunable to accommodate IO/compute capability of different hardware (e.g. due to diverse CPU/GPUs and DVFS). 
	
\end{myitemize}

\subsection{Instantiating Model Shards on Disk}
\label{sec:design:sharding:shard}
To address the challenges, our key idea is to combine two machine learning techniques -- dynamic \tfm{}~\cite{dynabert,hat} and dictionary-based quantization~\cite{deep-compression}, in a novel way.
We next describe details.

\begin{figure}[t]
	\centering
	\includegraphics[width=0.48\textwidth]{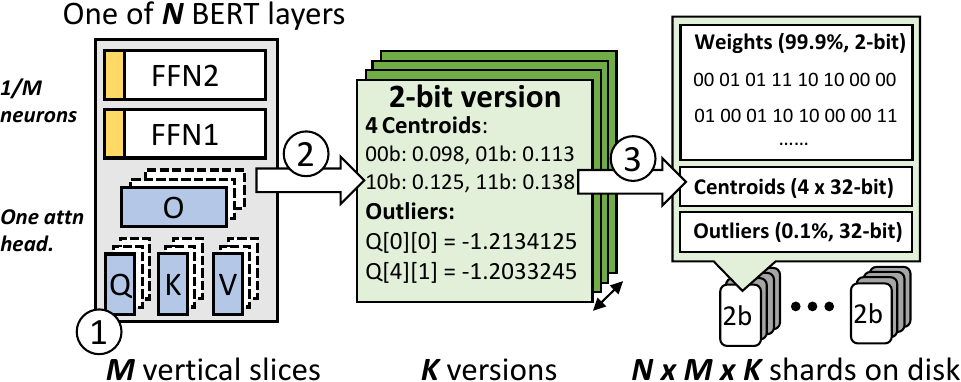}
	\caption{Instantiating $N\times M \times K$ model shards on disk. 
	The example shows a 2-bit shard.
	99.9\% of its weights are represented by 2-bit indexes pointing to $2^2$ centroids;
	the rest 0.1\% outliers are preserved as-is.
	}
	\label{fig:design:shards}
\end{figure}


\paragraph{First, vertical partitioning per layer}
The system adopts a pretrained \tfm{} model, which has already been fine-tuned on a downstream task. 

For each of the $N$ layers, the system partitions it into $M$ vertical slices, as shown in Figure~\ref{fig:design:shards}~(\circled{1}). 
By construction, each vertical slice is independent, constituting one attention head plus $1/M$ of FFN neurons of the layer; 
the partitioning is inspired by dynamic \tfm{}s~\cite{hat,dynabert}.
Table~\ref{tab:vertical_slice} shows the weight compositions of a vertical slice.
Each cell of the table describes the dimension of the weight matrix, where $d$ is the hidden state size, $M$ is the number of attention heads, and $d_{ff}$ is the number of FFN neurons;
a shard is therefore one of the $M$ equal slices of a layer. 
Doing so warrants model shards the same capability to extract linguistic features from inputs, as done by the attention mechanism: 
of an individual shard, its attention head obtains one independent representation of input tokens, which is projected into a higher feature dimension by FFN neurons~\cite{head-prune-acl19,bert-acl19};
jointly, multiple shards attend to information from different representation subspace at different positions~\cite{transformer}.
Therefore, an arbitrary subset of shards of a layer can be executed and still give meaningful results. 

\sys{} uses the \textit{submodel} to describe the \tfm{} model on shards, e.g. a  $n \times m$ submodel comprises $n$ layers, each layer having $m$ shards.
The number $m$ is the same across all layers, as mandated by the \tfm{} architecture~\cite{transformer}, which specifies each layer must have the same width (i.e. number of shards $m$) for aligning input/output features between layers.
Although it is possible for a shard to use 0s as dummy weights, \sys{} expects all $m$ shards to have concrete weights for a good accuracy.

\begin{table}
	\centering
	\caption{The weight composition of a shard. $M$ is number of attention heads. 
		The $M$ shards equally slices a \tfm{} layer, where each shard of the layer can be uniquely identified by its vertical slice index $i = 0 \dots M-1$.}
	\begin{tabular}{l|c|c|c}
		& Attn (Q,K,V,O)         & FFN1                        & \multicolumn{1}{l}{FFN2}     \\ 
		\hline
		Transformer Layer & $d\times d $           & $d_{ff} \times d$             & $d \times d_{ff}$              \\ 
		\hline
		Shard (vertical slice)      & $d \times \frac{d}{M}$ & $\frac{d_{ff}}{M} \times d$ & $d \times \frac{d_{ff}}{M}$  \\
		\hline
	\end{tabular}
	\label{tab:vertical_slice}
\end{table}

%



\paragraph{Second, quantization per shard}
\label{sec:design:sharding:quantize}
The system compresses each of the $N \times M$ shards into $K$ bitwidths versions (e.g. $K=2\dots 6$). 
\sys{} is the first to bring quantization to \textit{shard} granularity, whereas prior work only explores layer granularity~\cite{mixed-quantization-energy,hawq,haq}.
Doing so reduces IO/compute skewness and facilitates elastic IO, allowing \sys{} to prioritize IO resources at a much finer granularity, e.g. by allocating higher bitwidths to more important shards, and catering to IO/compute capability of diverse devices.



To compress, \sys{} uses Gaussian outlier-aware quantization~\cite{gobo}.
The key idea is to represent the vast majority of weights (e.g. 99.9\%) which follow a Gaussin distribution using $2^k$ floating point numbers (i.e. \textit{centroids});
doing so compresses the original 32-bit weights into $k$-bit indexes pointing to centroids, thus reducing the parameter size by $\frac{32}{k}$. 
For the very few \textit{outliers} (e.g. 0.1\%) which do not follow the Gaussian distribution, it preserves their weights as-is.
The process is shown in Figure~\ref{fig:design:shards}~(\circled{2}).
We will further describe the implementation details in \sect{impl}.

We choose it for two main reasons. 
1) It provides good compatibility between shards of different bitwidths, allowing \sys{} to tune their \bt{} individually per their importance and to assemble a \textit{mixed-bitwidth} submodel.
This is due to its lossy compression nature -- shards still preserve the original distribution of layer weights, albeit in different fidelities.
Hence they can work with each other seamlessly.
2) It does not need to fine-tune a model or require additional hardware support.
The quantization analyzes the weight distribution of the pretrained model and is not specific to network structures; 
it hence does not require fine-tuning, as opposed to fixed-point quantization~\cite{q8bert,bibert}. 
The resultant  \textit{mixed-\bt{}} submodel also differs from a traditional \textit{mixed-precision} network~\cite{haq,hawq,mixed-quantization-energy}, 
which requires DSP extensions for executing integer operations efficiently;
the extensions are often exclusive to microcontrollers on ARM devices, e.g. Cortex-M4~\cite{m4-dsp}.

Quantized shards are not meant to be used as-is.
Prior to use, \sys{} must decompress them, which is a mirror process of compression. 
\sys{} does so by substituting dictionary indexes with floating point centroids and outliers.
Therefore model shards quantization reduces IO but not computation (FLOPs) as the inference still executes on floating point numbers. 
\paragraph{Third, storing shards per version}
\sys{} stores each shard of every \bt{} on disk, in total $N\times M \times K$ shards (e.g. N=M=12, K=$2\dots6, 32$, where 32 is the uncompressed, full fidelity).
Each shard contains a weight matrix of the same dimensions listed in Table~\ref{tab:vertical_slice}. 
Instead of original FP32 weights, the weight matrix now stores K-bit indexes, which reduces its file size by $32/K \times$. 
Additional to the weight matrix, it stores centroids and outliers as dictionaries to look up during decompression, as illustrated by Figure~\ref{fig:design:shards}~(\circled{3}).
To load, it refers to individual on-disk shards by their original layer/vertical slice indexes and \bt{}s. 

\section{Pipeline planning}

\label{sec:design:planning}
\label{sec:planning}



\subsection{Overview}


\paragraph{Planning goals}
Towards maximizing the accuracy under a target latency T, \sys{} plans for two goals:
\begin{myitemize}
	\item 
	\textit{First, minimize pipeline bubbles.} 
	\sys{} attempts to utilize both IO and computation as much as possible: 
	by keeping IO always busy, it loads higher-\bt{} shards to improve submodel fidelity;
	by maxing out computation (FLOPs), it drives the inference towards a higher accuracy.
	
	
	
	\item 
	\textit{Second, prioritize \bt{}s on important shards.}
	As \tfm{} parameters exhibit clear redundancy, \sys{} allocates IO bandwidths with respect to shard importance, i.e. a shard is more important if it contributes more significantly to accuracy when being executed in higher \bt{}s.
%
\end{myitemize}

\paragraph{Two-stage planning}
Towards the goals, \sys{} conducts a two-stage planning:
1) Compute planning.
Based on measured computation delay of a layer, it proposes the largest submodel R' bound by T, which has the maximum FLOPs. 
2) IO planning.
It first assigns an \textit{accumulated IO budget} (AIB) to each layer of the submodel R' for tracking layerwise IO resources. 
To allocate and saturate the IO resources, \sys{} attempts to consume each layer's AIB.
Starting from most important shards, \sys{} assigns them a higher \bt{}, e.g. 6-bit; 
it does so iteratively for less important shards, until no AIB is left for each layer. 
We next describe details.



\subsection{Prerequisite: Offline Profiling}
\label{sec:design:planning:pre}
The following measurements are done ahead of time, off the inference execution path. 
\paragraph{Hardware capability}
\sys{}  measures the following hardware capabilities of a mobile device at installation time.
\begin{myitemize}
	\item 	
	IO delay $T_{io}(k)$ as a function of bitwidth $k$. 
	\sys{} measures the average disk access delay for loading one shard in $k$ \bt{}, where $k=2 \dots 6, 32$. 
	It only has to measure one shard per \bt{} because all others have same amount of parameters. 
	
	\item 
	Computation delay $T_{comp}(l, m, \textit{freq})$ as a function of $l$, the input sentence length, $m$,  the number of shards per layer (e.g. $m=3 \dots 12$), and $\textit{freq}$ as the current operating frequency of CPU/GPU.
	It fixes $l$ to be commonly used input lengths after padding (e.g. $l=128$). 
	It does a dry run for each $(l, m, \textit{freq})$ tuple on one \tfm{} layer. 
	It measures the average execution delay as the decompression delay of $m$ shards in 6-\bt{} and the execution delay of the \tfm{} layer composed by the $m$ shards.
	Although the decompression delay is strictly dependent on the shard \bt{}, the delay differences between individual \bt{}s are negligible in practice, e.g. $<1ms$;
	measuring 6-\bt{} shards further bounds the decompression delays, ensuring \sys{} always stays under the target latency. 

\end{myitemize}	

The delays can be recorded offline and replayed at run time because they are data-independent~\cite{driverlet,tinystack} and are shown deterministic~\cite{asymo}, w.r.t. the parameters $k, l, m, \text{and}\ \textit{freq}$.

\paragraph{Shard importance}
Intuitively, important shards have greater impacts on accuracy. 
Formally, \sys{} deems a shard more important than another if the shard increases the model accuracy more significantly \textit{as} they have higher fidelities. 
Specifically, \sys{} profiles shard importance as follows. 
It first sets the full 12x12 model (i.e. with 144 shards) to the lowest \bt{} (i.e. 2-bit), enumerates through each shard, and increases the shard \bt{} to the highest (i.e. 32-bit);
for each enumeration, it runs the resultant model on a dev set and profiles its accuracy.
The profiling therefore produces a table (e.g. with $12\times12=144$ entries), whose each entry records the model accuracy when the individual shard is at the highest \bt{} while all others are at the lowest \bt{}. 
\sys{} then sorts the table by model accuracy and obtains the list of ranked shard importance.

\begin{figure}[t]
	\begin{minipage}[t]{0.23\textwidth}
	\centering
	\includegraphics[width=1\textwidth]{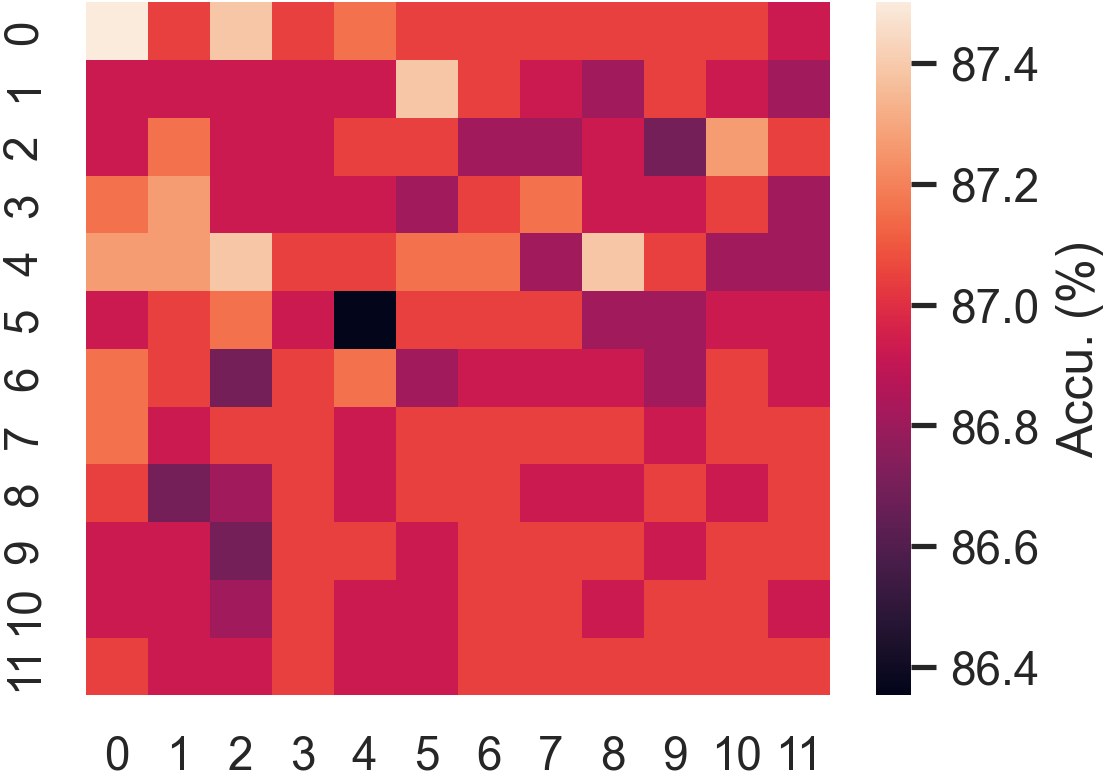}
	\subcaption{SST-2}	\label{fig:eval:importance:sst}			
\end{minipage}\hfill
\begin{minipage}[t]{0.23\textwidth}
	\centering
	\includegraphics[width=1\textwidth]{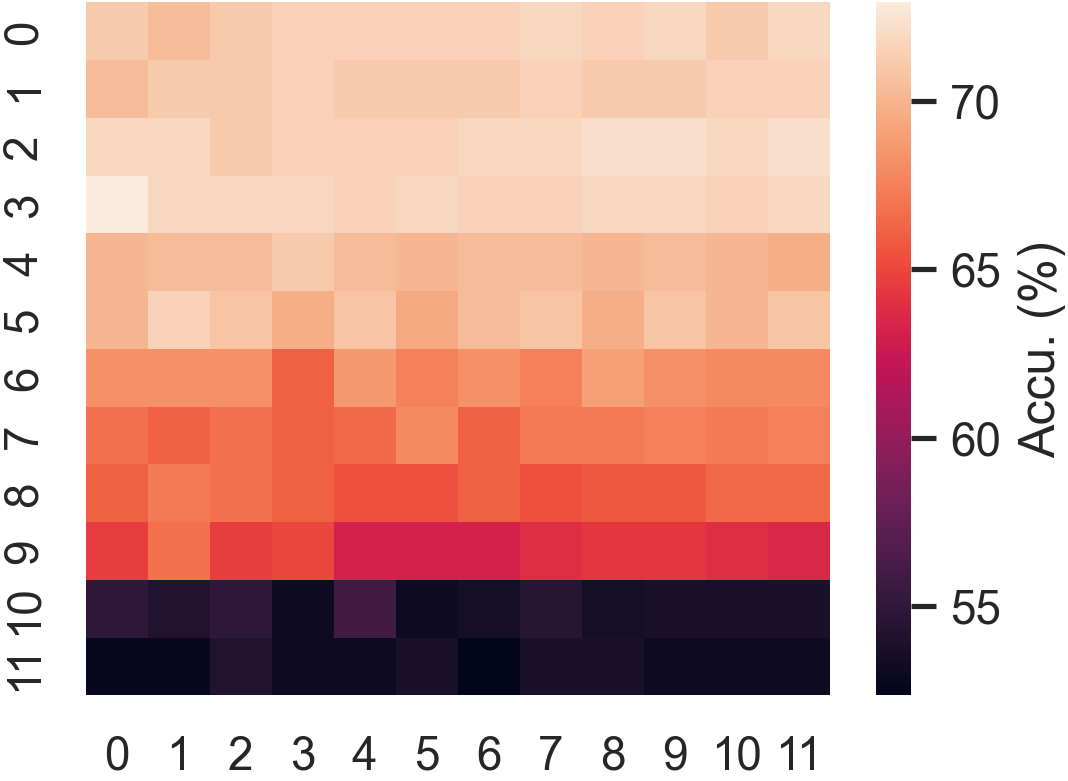}
	\subcaption{RTE}	\label{fig:eval:importance:rte}			
\end{minipage}\hfill
	\caption{Example shard profiles on SST-2 and RTE show distinct importance distribution.
	Each cell at (x, y) marks a shard; the lighter its color is, the more important the shard is (i.e. the higher accuracy during profiling).
	Y-axis: \tfm{} layer index, X-axis: vertical slice index.  }
	\label{fig:eval:importance}
\end{figure}

Notably, the profiling needs to be done for individual fine-tuned models, which have different weight distributions. 
Figure~\ref{fig:eval:importance:sst} and~\ref{fig:eval:importance:rte} shows the example of profiling results for models used in SST-2 and RTE respectively. 
As can be seen, shards of different models exemplify dissimilar importance distributions. 
For instance, important shards distribute more evenly throughout the layers of SST-2 model yet they are much more concentrated on bottom layers (i.e. layer 0-5) of RTE model.

\subsection{Compute Planning}
\label{sec:design:planning:compute}
Given a target latency T, \sys{} proposes a submodel sized by $n \times m$ for the incoming inference, which maximizes FLOPs.  

\paragraph{Key ideas}
In searching for the submodel size, \sys{} follows two principles: 
1) whenever possible, it always picks the submodel with \textit{most} number of shards, i.e. $n \times m$ is maximized; 
2) when two candidate submodels have similar number of shards, it prefers the deeper one, i.e. the candidate with a larger $n$.
As the \tfm{} attention heads within the same layer are known to be redundant~\cite{16heads}, it is wiser to incorporate more layers. 

To infer $(n, m)$, \sys{} enumerates through all possible pairs using the profiled $T_{comp}(l, m, freq)$;
the enumeration process has a constant complexity and is efficient. 
Since all inputs can be padded to a constant length (e.g. $l=128$), and $freq$ is often at peak during active inference, \sys{} only needs to enumerate in total 144 pairs in practice. 
For each T, the enumeration therefore deterministically gives a submodel of $(n \times m)$ which is both largest and deepest.

\subsection{IO Planning}
\label{sec:design:planning:io}
In this stage, \sys{} selects the \bt{}s for individual shards of the $(n\times m)$ submodel. 
Without stalling the pipeline, it seeks those that maximize accuracy. 

%
%
%
%


\subsubsection{Problem Formulation}
Given the deadline $T$, $n \times m$ submodel $R$ determined by compute planning, and the preload buffer $S$, \sys{} plans for a shard configuration $S'$ to load during computation, s.t. 1) loading $S'$ does not stall the pipeline, and 2) $R=S+S'$ achieves maximum accuracy.




\subsubsection{Accumulated IO Budgets}
To ensure the planning $S'$ does not stall the pipline, \sys{} uses \textit{Accumulated IO Budgets} (AIBs) to track fine-grained, per-layer available IO bandwidth. 

\paragraph{Key ideas}
To quantify AIBs, our observation is that the pipeline does not stall \textit{iff} before executing one layer, all shards of the current \textbf{and} prior layers are already loaded.
We hence define AIBs as follows: 

\begin{definition}[Accumulated IO Budgets]
The AIB(k) of $k^{th}$ layer is the available IO time the layer can leverage to load all shards from $0\dots k$ layers, written as 
$AIB(k) = AIB(k-1) + T_{comp}(k-1)$, where $T_{comp}(k-1)$ is the computation delay of the $(k-1)^{th}$ layer.
\end{definition}


The recursive definition (i.e. hence \textit{accumulated}) encodes the data dependency between pipeline layers: each layer crucially depends on previous layers' available IO budgets and computation delays for overlapping the loading of its own shards.
As of the very first layer, its AIB is set as the IO delay to fill the preload buffer $S$, considered as ``bonus IO''. 
For instance, the AIB of the second layer is the AIB \textit{plus} the computation delay of the first layer, i.e. $AIB(1)=AIB(0)+T_{comp}(0)$.
With the above definition, \sys{} checks AIBs of all layers: as long as they are non-negative, \sys{} knows each layer still has IO time remaining and the pipeline does not stall, and deems the planning \textit{valid}.  

\paragraph{How to use}
Upon each planning, \sys{} initializes AIBs for all layers as follows.  
It first sets AIB(0) to be the IO delay to fill the preload buffer as described before.
Next, \sys{} sets subsequent AIBs recursively using the above definition, e.g. $AIB(1)=AIB(0) + T_{comp}, AIB(2) = AIB(0) + 2\times T_{comp}, AIB(3)=AIB(0) + 3\times T_{comp}$.  
Note that since layers have an identical structure, \sys{} uses a constant $T_{comp}$ across all layers. 

When \sys{} selects a shard at $k$-th layer, it deducts the shard IO from AIBs of $k$-th as well as all subsequent layers.
This is because loading such shards only affect \textit{yet-to-be-executed} layers but not the already executed ones. 
At the end of selection, \sys{} checks all AIBs to see if they are non-negative. 
If so, \sys{} deems the planning $S'$ valid, otherwise rejects it.

\begin{figure}[h]
	\includegraphics[width=0.48\textwidth]{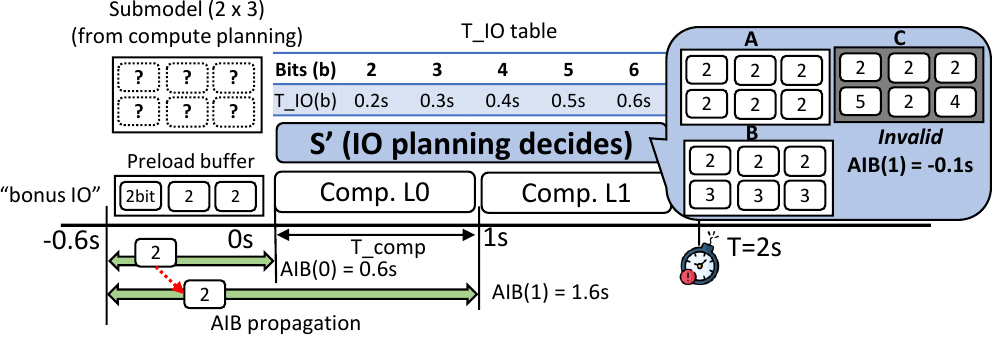}
	\caption{A mini example of AIB tracking the layerwise IO budgets.}
	\label{fig:design:aib}
\end{figure}

\paragraph{Example}
Figure~\ref{fig:design:aib} shows a mini example of using AIBs to check the validity of $S'$, where it plans for a 2x3 submodel, targeting a 2s deadline with $T_{comp}=1s$.
The engine initializes AIBs recursively from L0, whose $AIB(0)=0.6s$ due to the three 2-bit shards in $S$.
To plan, the engine first fills $S'$ with $S$, deducting $0.6s$ from both $AIB(0)$ and $AIB(1)$ because all shards in $S$ are in L0.
Since only L1 has spare AIB, the engine can only select shards for it.
We show three execution plan candidates A, B, and C.
In this case, both candidates A and B are valid because their AIBs are non-negative, meaning loading them does not stall computation L1. 
Yet, C is invalid, because $AIB(1)=-0.1s$, violating the constraint and stalling the pipeline.


\subsubsection{Selecting Optimal Shard Versions}
For each $T$, there exist an enormous number of execution plans. 
The goal is to select an optimal configuration $S'$, which 1) is valid, and 2) maximizes accuracy.
For instance, both A and B in Figure~\ref{fig:design:aib} are valid, but which has the maximum accuracy?

\paragraph{Key idea} 
To ensure validity, \sys{} respects the key invariant $AIB(k) \geq 0$ for each allocation attempt on layer $k$.
To maximize accuracy, our key idea is to first uniformly increase \bt{}s for all shards, then with the rest AIBs it greedily and iteratively allocates highest possible \bt{}s to individual shards guided by shard importance.
By doing so, we build an information passageway for most important shards, allowing their maximum activations to be preserved in as high fidelity as possible. 

The allocation process comprises two passes as follows.
In the first pass, \sys{} picks a uniform \bt{} for all unallocated shards in the submodel, i.e. those not in preload buffer.
To do so, it enumerates from lowest \bt{} (i.e. 2-bit) and selects the highest \bt{}s while AIBs still satisfy the invariant.
Notably, it fills a submodel layer with the shards from the same original layer and does not mix up shards across layers, due to quantization preserves intra-layer weight distribution (\S\ref{sec:design:sharding:quantize}).
If AIBs cannot even support 2-bit shards, e.g. due to $T$ and/or preload buffer $S$ too small, \sys{} still selects them as they are necessary for execution but aborts further allocation.
In the second pass, \sys{} iteratively upgrades the \bt{}s of individual shards to full 32 \bt{} guided by the shard importance profiled in \S\ref{sec:design:planning:pre}, until all AIBs are consumed. 

The allocation result is an optimal execution plan which instantiates the submodel with individual shard configurations, and is ready to be executed by the IO/compute pipeline.
\subsection{Submodel Execution}
\label{sec:design:planning:pipeline}


\sys{} executes the plan (i.e. the $n\times m$ submodel with selected shards) in a layerwise, pipelined manner from layer 0 to layer n-1.
While conceptually it is possible to pipeline shard computation within the same layer, \sys{} does not do so due to limited benefits -- within a layer there exists data dependency between the FFNs and attention module. 

\sys{} executes both IO and computation as fast as possible;
it does not reorder the loading of individual shards in order to meet data dependency between execution, because by design AIBs have already ensured so.  
To compute, \sys{} decompresses the shards into the working buffer using the dictionaries stored along with them;
the working buffer is enough to hold one layer of FP32 weights and shared by all layers during their ongoing execution. 
After execution, \sys{} evicts loaded shards from top to bottom layers until preload buffer is filled. 
It does so because shards at bottom layers (i.e. closer to input) are needed early during inference.
Preserving as many of them as possible avoids compulsory pipeline stalls in early stages.
\section{Implementation}
\label{sec:impl}

We implement \sys{} in 1K SLOC (Python: 800, C: 200) based on PyTorch v1.11~\cite{torch-1.11} and sklearn v0.23.2~\cite{sklearn-23.2}, atop two commodity SoCs listed in Table~\ref{tab:platform}. 

We preprocess the pretrained DynaBERT~\cite{dynabert} models. 
We choose them because they are easily accessible and well documented. 
We preprocess the model as follows. 
To quantize a model into k \bt{}, we first partition the model by layers and gathers all weights of the layer into a large flat 1D array.
We then fit the 1D array into a Gaussian distribution using \code{GaussianMixture} with one mixture component from \code{sklearn.mixture} for detecting outliers.
Based on the fitted distribution, we calculate the log likelihood of each sample in the 1D weight array.
Following~\cite{gobo} we also use -4 as the threshold -- if the weight's log likelihood is below the threshold, we deem it as an outlier and records its array index;
in our experiments, a model only has 0.14-0.17\% outliers, which are an extremely small portion. 
For non-outliers which are the vast majority, we sort them based on their values and divided them into $2^k$ clusters with equal population.
We calculate the arithmetic mean of each cluster as one centroid for representing all weights of the cluster.
With such, we extract shards from the layer based on their weight composition in Table~\ref{tab:vertical_slice} and massively substitutes their weights with k-bit indexes to centroids;
for bit alignment, we represent outliers also as k-bit integers but bookkeep their original weights and offsets in the shard. 
We repeat the process for each layer and for each $k=2\dots 6$, which takes a few minutes per \bt{}.
We co-locate disk blocks of shards from the same layer for access locality.
To measure shard importance, we use dev set from the respective GLUE benchmark on which the model is fine-tuned.

Implementing the layerwise pipeline is straightforward, by intercepting the forwarding function at each BERT layer and using asynchronous IO for loading shards.
Yet, we have discovered Python has a poor support for controlling concurrency at fine granularity (e.g. via low-level thread abstraction), which introduces artificial delays to shard decompression. 
Therefore we implement the decompression in separate 200 SLOC of C code using OpenMP~\cite{dagum1998openmp}, which concurrently substitutes the low-bit integers back to FP32 centroids using all available cores of our SoCs;
we expect the decompression to be further accelerated with GPU, but leave it as future work. 

For miscellaneous parameters of a layer which are not part of shards, i.e. layer normalization (layernorm) and biases, we keep them in memory in full fidelity because their sizes are small, e.g. tens of KB per layer.
\begin{table}[t]
	\centering
	\caption{
		Platforms in evaluation.
		Benchmarks run on Odroid's CPU (its GPU lacks Pytorch support) and Jetson's GPU. 
	}
	\includegraphics[width=0.4\textwidth{}]{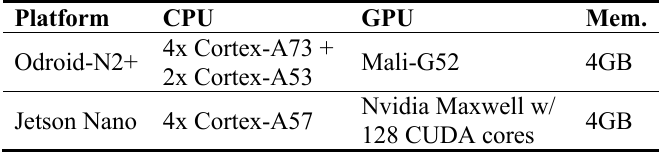}
	\label{tab:platform}
\end{table}

\vspace{2em}
\section{Evaluation}
\label{sec:eval}
We answer the following questions: 
\begin{myenumerate}
	\item 
	Can \sys{} achieve competitive accuracy under time and memory constraints? (\S\ref{sec:eval:deadline}) 
	
	\item 
	How much do \sys{}'s key designs contribute to its performance? (\S\ref{sec:eval:micro})
	
	\item 
	How do \sys{}'s benefits change with available time and memory? (\S\ref{sec:eval:sensitivity})
\end{myenumerate}



\subsection{Methodology}
\label{sec:eval:method}

\paragraph{Setup and metrics}
Table~\ref{tab:platform} summarizes our test platforms, which are commodity SoCs.
We choose them to evaluate \sys{} on both CPU and GPU.
Based on user satisfaction of NLP inference delays on mobile devices~\cite{nlp-delay-study}, we set T=150, 200, and 400ms. 
Prior work reported that beyond 400ms user satisfaction greatly drops~\cite{nlp-delay-study}.
With T under 100ms, all comparisons including \sys{} show low accuracy -- there is not enough compute bandwidth. 
This is a limit in our test hardware, which shall mitigate on faster CPU/GPU. 

Table~\ref{tab:benchmarks} summarizes our benchmarks and metrics. 
We diversify them to include each category of GLUE benchmarks~\cite{glue-benchmark}, which span a broad range of NLP use cases on mobile devices. 

\begin{table}[t]
	\centering
	\caption{
		GLUE benchmarks~\cite{glue-benchmark} used in evaluation.
	}
	\includegraphics[width=0.48\textwidth{}]{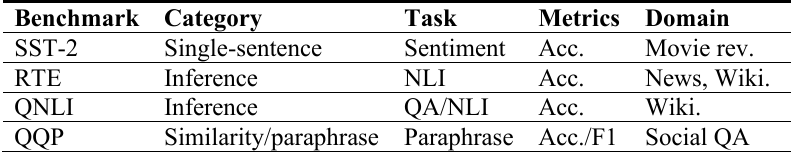}
	\label{tab:benchmarks}
\end{table}

\paragraph{Comparisons} 
We consider two NLP models. 
(1) DistilBERT~\cite{distilbert}, the outcome of knowledge distillation from BERT. 
Due to its high popularity on mobile, we use its accuracy as our references and call it \textit{gold accuracy}.
Yet, DistilBERT has fixed depths/widths (6 layers x 12 heads) and thus cannot adapt to different target latencies. 
(2) DynaBERT~\cite{dynabert}, which is derived from BERT (12 layers x 12 heads), allowing execution of a submodel to meet the target latency. 

Based on DynaBERT, we design the following competitive baselines as summarized in Table~\ref{tab:baselines}. 
\begin{table}[t]
	\centering
	\caption{
		Baselines for evaluation and their positions in the design space.
	}
	\includegraphics[width=0.48\textwidth{}]{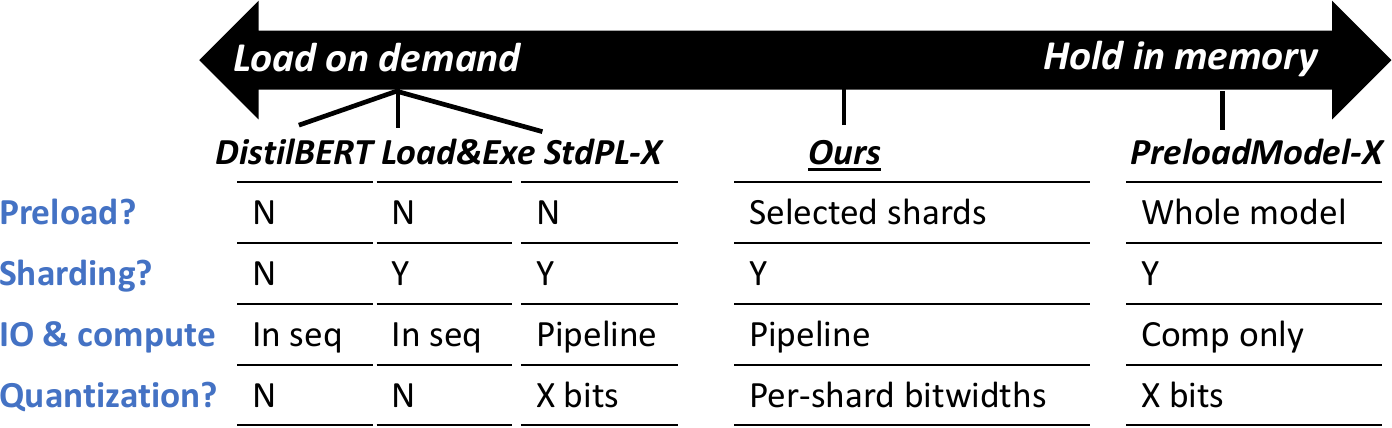}
	\label{tab:baselines}
\end{table}


\begin{myitemize}
	\item \textit{Load\&Exec}: 
	It loads model as a whole and executes it.	
	It chooses the best submodel so the sum of IO and execution delays is closest to the target latency, using the algorithm described in Section~\ref{sec:design:planning:compute}. 
	Model parameters are not quantized (32 bits). 

	
	\item \textit{Standard pipelining (StdPL-X)}: 
	It executes a layerwise pipeline, overlapping IO and computation. 
	It chooses the best submodel so that the total pipeline delay stays under the target latency. 
	We further augment it with quantization. 
	All parameters in a model have the same \bt{} X.

	\item \textit{PreloadModel-X}: 
	The whole model is already in memory and no IO is required. 
	It chooses the best submodel so that the total computation delay stays under the target latency. 
	We augment it with quantization;
	all parameters have the same \bt{} X.
\end{myitemize}

We choose X=6 as the highest quantization \bt{}, as further increasing the \bt{} has little accuracy improvement. 

\begin{table*}[h!]
	\centering
	\caption{Model execution accuracies; given target latencies, ours are the best or the closest to the best. 	
		|S|: preload buffer size. 
		Gold accuracy from DistilBERT~\cite{distilbert}, 
		which exceed all target latencies.
		End-to-end DistilBERT execution delays: 3.7s on Odroid, of which IO=3.1s; 
		3.36s on Jetson, of which IO=3.0s. 
	}
	\includegraphics[width=1\textwidth{}]{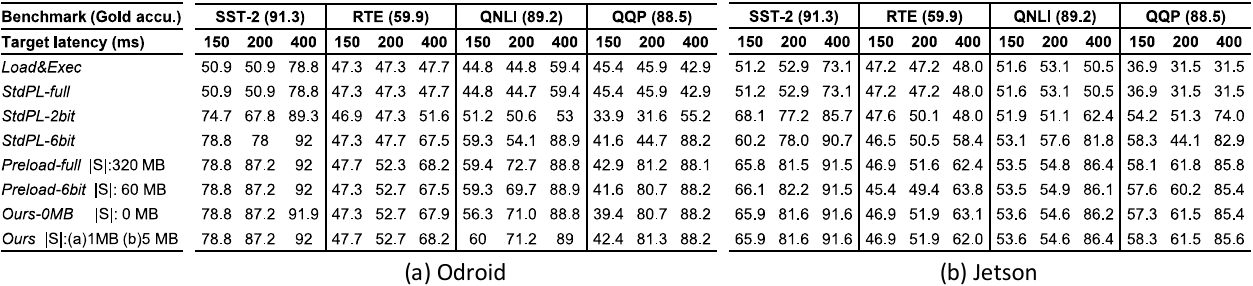}
	\label{tab:accu}
\end{table*}

\subsection{End-to-End Results}
\label{sec:eval:deadline}



\sys{} achieves comparable accuracies to \textit{gold} under target latencies (T) of a few hundred ms. 
Across all benchmarks and latencies, \sys{} accuracy is on average 7.1 percentage point (pp) higher than that of baselines, which is significant.

Compared to preloading the whole model, \sys{} reduces memory consumption by 1-2 orders of magnitude while seeing 0.16 pp higher accuracy averaged across all latencies and benchmarks; 
compared to loading the model on demand, \sys{} improves the accuracy by 14 pp at the cost of preload memory of no more than 5 MBs. 

\begin{figure}[t]
	\begin{minipage}[t]{0.48\textwidth}
		\centering
		\includegraphics[width=0.48\textwidth]{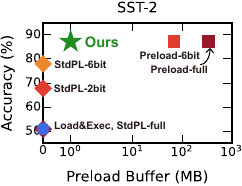}
		\includegraphics[width=0.48\textwidth]{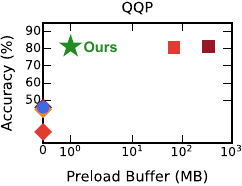}
		\subcaption{Odroid}	
	\end{minipage}\hfill
	\begin{minipage}[t]{0.48\textwidth}
	\centering
	\includegraphics[width=0.48\textwidth]{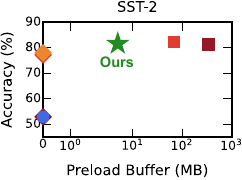}
	\includegraphics[width=0.48\textwidth]{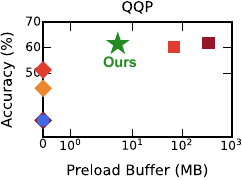}
	\subcaption{Jetson}
	\label{fig:eval:memory:jetson}
	\end{minipage}\hfill
	\caption{
	\sys{}'s accuracy is significantly higher than \textit{Load\&Exec} and \textit{StdPL}, and is similar/higher compared to \textit{PreloadModel} albeit using 1-2 orders of magnitude smaller memory. 
	T=200ms. 
	Full data and benchmarks in Table~\ref{tab:accu}.}
	\label{fig:eval:memory}
\end{figure}

Figure~\ref{fig:eval:memory} zooms in accuracy/memory tradeoffs under $T=200ms$ of SST and QQP benchmarks. 
Note that we use log scale in X-axis (memory consumption) due to its large span.
\sys{} uses 204$\times$ lower memory than \textit{PreloadModel-full} while having less than 1\% average accuracy loss.
Even when compared with the quantized version (i.e. \textit{PreloadModel-6bit}), \sys{} uses on average 41$\times$ smaller memory to achieve the same accuracy.

\paragraph{Accuracy}
\sys{}'s accuracy matches those of DistilBERT.
Given a target latency T, \sys{} achieves consistent and significant accuracy gain over baselines.
Table~\ref{tab:accu} shows the full view.
On Odroid, \sys{} (Ours) increases average accuracy by \sloppy{21.05/21.05/17.13/5.83} pp compared with \textit{Load\&Exec}/\textit{StdPL-full}/\textit{StdPL-2bit}/\textit{StdPL-6bit}, respectively. 
On Jetson, \sys{} increases average accuracy by 18.77/18.77/6.53/3.15 pp compared with \textit{Load\&Exec}/\textit{StdPL-full}/\textit{StdPL-2bit}/\textit{StdPL-6bit}, respectively. 
Notably, \sys{}'s benefit is game-changing compared with \textit{Load\&Exec} and \textit{StdPL-full}. 
They are barely usable under low latency (T$\leq$200ms). 
\paragraph{Memory consumptions}
Compared with preloading the whole model, \sys{} reduces memory consumption significantly and consistently, by 122$\times$ on average. 
This is because the \textit{PrelodModel} baselines hold the whole 12x12 model in memory. 
By  comparison, \sys{} only needs preload memory of 1MB/5MB on Odroid and Jetson respectively, which is sufficient to hold shards of the first model layer and warms up the pipeline execution. 
\paragraph{Storage \& energy overhead}
For a model, \sys{} only requires 215 MB disk space to store five fidelity versions of \{2,3,4,5,6\} bits, in addition to the full model (in 32 bits) of 418 MB. 
This storage overhead is minor given that today's smartphone has tens or hundreds GB of storage.

For a given latency, we expect \sys{} to consume notably more energy than low-accuracy baselines (e.g. \textit{Load\&Exec}, \textit{StdPL-full}), as \sys{} has higher resource utilization to achieve higher accuracy. 
Compared to similar-accuracy, high-memory baselines (i.e. \textit{PreloadModel-full}), we expect \sys{} to consume moderately but not significantly more energy. 
First, the major energy consumer is active compute (FLOPs); similar accuracies indicate similar FLOPs. 
Second, although \sys{} adds IO activities, the contribution to the system power is marginal because the whole SoC is already in high power states. 


\begin{table}[h]
		\caption{Sizes (depth$\times$width) of submodels selected under different target latencies. A large submodel means more FLOPs executed, suggesting a higher accuracy. 
		\sys{} is able to run the largest submodel.
		}
		\includegraphics[width=0.48\textwidth{}]{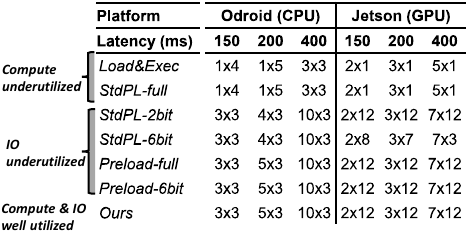}
		\label{tab:network-confs}
\end{table}

\subsection{Significance of Key Designs}
\label{sec:eval:micro}



\begin{figure}[h]
	\includegraphics[width=0.45\textwidth]{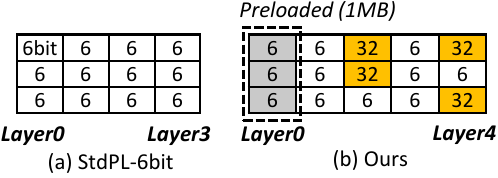}
	\caption{A comparison between submodels executed by \textit{Ours} and \textit{StdPL-6bit}. Benchmark: SST-2 on Odroid. T=200ms.
	\textit{Ours} runs a larger submodel and higher FLOPs, resulting in 9.2 pp higher accuracy.}
	\label{fig:submodel}
\end{figure}

\paragraph{Submodel configuration} 
Within a given latency, the result accuracy hinges on total FLOPs executed, which depends on the size of executed submodel. 
Our results show that \sys{} dynamically adjusts submodel sizes towards the maximum FLOPs. 
Table~\ref{tab:network-confs} shows the details. 
Estimated by comparing submodel sizes: 
our FLOPs is as high as that of \textit{PreloadModel}, which however consumes 1-2 orders of magnitude more memory; 
our FLOPs is 7$\times$ higher compared with \textit{Load\&Exec} and \textit{StdPL-full}, for which the IO blocks computation most of the time; 
our FLOPs is 1.3$\times$ higher than that of \textit{StdPL-2/6bit}, two strong baselines that increase FLOPs through IO/compute parallelism and quantization as us;
at lower T (e.g. $T \leq 200ms$), their IO delays of loading the first layer may block computation, resulting in a smaller model. 
Figure~\ref{fig:submodel} shows such an example. 
Thanks to a small preload buffer, our executed submodel has 1.25$\times$ higher FLOPs (i.e. it has one extra layer), which leads to 9.2 percentage point (pp) higher accuracy. 


Table~\ref{tab:network-confs} also shows that our system adjusts submodels according to platform hardware. 
Specifically, our system assembles shallow/wide submodels on Jetson (GPU) as opposed to deeper/narrower submodels on Odroid (CPU). 
The reason is GPU's lack of proportionality on Transformer shards, 
e.g. executing a layer of 12 shards is only 0.7\% longer than a layer of 3 shards.
The root cause is that GPU is optimized for batch workload; 
it pays a fixed, significant cost even in executing a fraction of a transformer layer and for one input example, which is the case of interactive NLP. 


%

%
%


\paragraph{Elastic pipelining}
\sys{}'s per-shard bitwidths contribute to its accuracy significantly. 
By contrast, one fixed bitwidth for all shards in a model is too rigid, resulting in pipeline bubbles. 
With a full bitwidth of 32 bits (\textit{StdPL-full}), IO takes long and stalls the computation (19.9 pp lower accuracy than \sys{}); 
with a lower bitwidth (\textit{StdPL-\{2,6\}bit}), IO bandwidth is left underutilized (8.2 pp lower accuracy than \sys{}). 
Any fixed bitwidth between 6 and 32 bits does not help either (\sect{eval:method}). 
Unlike them, \sys{} well utilizes both compute and IO through its two-stage planning (\S\ref{sec:design:planning}).

\paragraph{Preload buffers} show a clear benefit as shown in Table~\ref{tab:accu}. 
By using a small preload buffer of a few MBs, 
\sys{} achieves a noticeable and consistent accuracy gain compared to not using the preload buffer (\textit{Ours-0MB}). 
The benefit is most pronounced on QNLI and QQP among the benchmarks, increasing accuracy by up to 3.7 percent point (Odroid). 
Section~\ref{sec:eval:sensitivity} will present a sensitivity analysis regarding its size. 

\paragraph{Shard importance} 
\sys{} allocates its IO budgets to the most important shards. 
The accuracy benefit is most pronounced in a small/median submodel where most shards have low to medium bitwidths. 

\textit{Case study.}
We demonstrate the efficacy through a differential analysis. 
Table~\ref{tab:eval:shard-selection} shows an intermediate state of planning: a 5x3 submodel comprising all 2-bit shards.
Now the planner is awarded additional IO budgets, e.g. from enlargement of the preload buffer, with which the planner will increase some shards' bitwidths to 6 bits. 
We compare two strategies: (1) randomly pick shards; (2) pick shards in their importance order (as in \sys{}). 
Despite the same IO budget is spent, \sys{} shows higher accuracy by up to 23.1 percent point (8.19 percent point on average) across all benchmarks.

\begin{table}[t]
	\centering
	\caption{Model accuracies resultant from allocating additional IO budget within a 5x3 submodel of 2-bit shards. Our method shows much higher accuracies than random shard selection.}
	\includegraphics[width=0.48\textwidth{}]{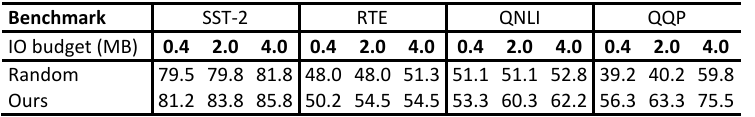}
	\label{tab:eval:shard-selection}
\end{table}


%
%

%


\subsection{Sensitivity Analysis}
\label{sec:eval:sensitivity}

We examine how \sys{}'s benefit changes as resource amounts. 

\paragraph{Target latencies}
A more relaxed target latency allows \sys{} to deliver more FLOPs and execute a deeper submodel, suggesting a higher accuracy. 
Yet, an NLP model's accuracy sees diminishing return as its depth continues to grow, as shown in prior work~\cite{layerdrop, dynabert}; 
as a result, \sys{}'s benefit diminishes as the target latency is further relaxed.  
Specifically, on Odroid (CPU) \sys{} has most significant advantage over baselines (7.7 pp higher accuracy) when target latencies are below 200 ms; 
in such cases, a feasible submodel has fewer than 10 layers. 
On Jetson (GPU) \sys{} has most significant advantage when target latencies are below 400 ms and a feasible submodel has fewer than 7 layers. 
When the target latency grows beyond such ranges, \sys{}'s benefits gradually reduce.

\paragraph{Preload buffer size}
Its significance hinges on the relative speeds of computation (which consumes model parameters) and IO (which loads the parameters), because the buffer bridges the speed gap of the two. 
When the computation is much faster than IO, an increase in the buffer size will result in large accuracy gain, and vice versa. 

On our platforms, \sys{} shows a noticeable and consistent accuracy gain over baselines by using a preload buffer of a few MBs. 
Since at current preload buffer size \sys{} has already reached best accuracy (i.e. same as \textit{PreloadModel-full}), further increasing the buffer size does not boost the accuracy proportionally. 
We expect that with faster compute (e.g. neural accelerators), the preload buffer takes in a greater role.
The reason is, when execution become faster and can only overlap with loading of low-fidelity shards (e.g. 2 bits), a few high-fidelity shards provided by preload buffer can significantly boost the accuracy.
Such a case is shown in Table~\ref{tab:eval:shard-selection}, as preload buffer sizes increase from 0.4 to 4.0 MB, the accuracy increase by 19.2 pp.
\balance
\section{Related work}

Our system is related to a wide range of ML and systems techniques.
We next discuss the similarities and differences.
\paragraph{Model compression} is a common technique for reducing model size (IO), facilitating faster loading;
it includes model structure~\cite{layerdrop,distilbert} and feature pruning~\cite{mobilebert}, and quantization which reduces full-precisions (32bit) parameters into low-bit (e.g. 2bit) representations~\cite{bibert,binarybert,q8bert,deep-compression}.  
We use quantization to compress the model; 
differently, we scale compression ratios to runtime IO by instantiating multiple compressed versions. 
Automated quantization searches for optimal bit-widths of a NN in the offline, often on a per layer basis~\cite{haq,hawq,emq}.
HAQ~\cite{haq} adopts the reinforcement learning to find the best mixed precision for each layer, similar with our multiple versions of shards.
Compared with them, we do not need any fine-tuning, which is time-consuming and we must make fine-grained decisions (i.e. per-shard) at run time.

\paragraph{Dynamic configuration of DNNs} 
changes model widths and/or depths in order to suit resource constraints~\cite{dynabert,edgebert,layerdrop,hat,nestdnn}.
EdgeBERT~\cite{edgebert} improves NLP energy efficiency under 
target latencies via early exit. 
NestDNN~\cite{nestdnn} hosts one multi-capacity model on device and switches across submodels depending on available resources. 
Assuming the whole model always held in memory, 
these systems miss the opportunities of pipelined IO/compute and therefore incur high memory cost when applied to NLP. 
Similar to them, we configure the NLP model architecture dynamically. 
Unlike them, we address the challenge of loading large models through pipelining. 
Furthermore, our configuration is on the basis of individual shards and adapts  to both memory and latency constraints. 

\paragraph{Pipeline parallelism for ML}
Pipelining has been extensively used to accelerate ML. 
Prior work mainly uses it to scale out ML to multiple machines (overcome limit of single machine resource). 
Notably for training, PP is used to partition a model or training data over a cluster of machines~\cite{gpipe} for maximizing hardware utilization by minimizing pipeline stalls using micro/minibatches~\cite{pipedream}, exploiting hardware heterogeneity~\cite{edgepipe}, or by adapting pipeline depths on the fly~\cite{pipetransformer}.
We share a similar goal of maximizing pipeline utilization and minimizing bubbles.  
Unlike that they focus on a pipeline of computations (forward/backward passes of different inputs) or network/computation, our pipeline consists of disk IO tasks and computation. 
Our approach towards high efficiency is through adjusting IO workloads of model shards to the computation. 

\section{Concluding remarks}

We present STI, a novel system for speedy transformer inference on mobile devices. 
STI contributes two novel techniques: model sharding and elastic pipeline planning with a preload buffer.
The former allows \sys{} to tune model parameters at fine granularities in a resource-elastic fashion.
The latter facilitates \sys{} for maximizing IO/compute utilization on most important parts of the model.
With them, STI reduces memory consumption by 1-2 orders of magnitude while delivering high accuracies under a practical range of target latencies.


\section*{Acknowledgment}
The authors were supported in part by NSF awards \#2128725, \#1919197, \#2106893, and Virginia's Commonwealth Cyber Initiative. 
The authors thank the anonymous reviewers for their
insightful feedback.




\bibliographystyle{ACM-Reference-Format}
\bibliography{bib/abr-short,bib/xzl,bib/hongyu,bib/misc,bib/book,bib/security,bib/iot,bib/datacentric,bib/hp,bib/transkernel,bib/ml-edge,bib/secureGPU,bib/TrustZone,bib/lwg,bib/sysml,bib/mm}



\end{document}